\documentclass[10pt,twocolumn]{article}

\usepackage[
    a4paper,
    top=0.70in,
    bottom=0.75in,
    left=0.70in,
    right=0.70in,
    columnsep=0.25in
]{geometry}
\usepackage{abstract}
\usepackage{tabularx}
\usepackage[utf8]{inputenc}
\usepackage[T1]{fontenc}
\usepackage{lmodern}
\usepackage{hyperref}
\usepackage{url}
\usepackage{booktabs}
\usepackage{amsmath}
\usepackage{amssymb}
\usepackage{amsfonts}
\usepackage{nicefrac}
\usepackage{microtype}
\usepackage{graphicx}
\usepackage{orcidlink}
\usepackage{array}
\usepackage{enumitem}
\usepackage{caption}
\usepackage{subcaption}
\usepackage{dblfloatfix} 
\usepackage{ragged2e}
\usepackage[table]{xcolor}

\newcolumntype{L}{>{\RaggedRight\arraybackslash}X}
\newcolumntype{C}[1]{%
    >{\Centering\arraybackslash}p{#1}}

\newcommand{\modelid}[1]{%
    {\ttfamily\footnotesize #1}}

\hypersetup{colorlinks=true, linkcolor=blue, citecolor=blue, urlcolor=blue}
\graphicspath{{./images/}}

\newcommand{\keywords}[1]{%
    \par\medskip
    \noindent\textbf{Keywords: }%
    \begingroup\def\and{\unskip, }#1\endgroup
    \par
}

\definecolor{skyblue}{RGB}{0, 191, 245} 
\hypersetup{
    colorlinks=true,       
    citecolor=skyblue,        
    linkcolor=red,         
    urlcolor=magenta       
}

\title{ChaosProbe: A Neurochaotic Lens on Frozen Transformer Input-Embedding Spaces}

\author{
    \textbf{Kunal Kumar Pant}\,
    \orcidlink{0009-0002-2748-8096}%
    \textsuperscript{\ensuremath{\dagger}}\\[0.3em]
    {\small Department of Electronics and}\\
    {\small Communication Engineering (ECE)}\\
    {\small Amrita Vishwa Vidyapeetham}\\
    {\small Kollam, Kerala, India}
    \and
    \textbf{Nithin Nagaraj}\,
    \orcidlink{0000-0003-0097-4131}%
    \textsuperscript{\ensuremath{\ddagger}}\\[0.3em]
    {\small Complex Systems Programme}\\
    {\small National Institute of Advanced Studies (NIAS)}\\
    {\small IISc Campus, Bengaluru}\\
    {\small Karnataka, India}
}

\date{}

\begin{document}

\twocolumn[
\begin{@twocolumnfalse}
\maketitle

\begin{onecolabstract}
Transformer models are most often understood through what they do: their benchmark performance, generation quality, or behavior on downstream tasks. Yet frozen transformer input-embedding spaces may also be examined through their responses to a controlled deterministic probe before contextual computation or task-specific adaptation. Guided by this response-based view, we introduce \emph{ChaosProbe}, a deterministic neurochaos-inspired method for constructing response-based fingerprints of frozen transformer input-embedding spaces. For each prompt-level embedding matrix, ChaosProbe applies a chaotic trajectory-based transformation and summarizes its Firing Rate and Entropy channel responses with complementary representation-level measures, producing a fixed-length signature for each model. In a bounded proof-of-concept study of $80$ neutral prompts and four pretrained models---GPT-2, DistilGPT2, BERT-base-uncased, and RoBERTa-base---Pearson correlation, Spearman correlation, and cosine similarity each recover all four same-family nearest-neighbor assignments and both expected mutual family pairs. Euclidean distance recovers three of the four assignments and
one of the two mutual family pairs. Paired bootstrap resampling supports the stability of the Pearson and Spearman
pairings over the observed prompt set, and signature-validity checks show that constant or collapsed responses do not dominate the reported fingerprints. These results provide a cohort-dependent proof of concept that deterministic neurochaotic response signatures can expose broad structure among frozen transformer input-embedding spaces.
\end{onecolabstract}


\vspace{1em}
\end{@twocolumnfalse}
]

\begingroup
\renewcommand{\thefootnote}{\fnsymbol{footnote}}

\footnotetext[2]{%
    \href{mailto:kunal.kpant@gmail.com}
         {\texttt{kunal.kpant@gmail.com}}%
    \hfill
    \textsuperscript{\ensuremath{\ddagger}}\,%
    \href{mailto:nithin@nias.res.in}
         {\texttt{nithin@nias.res.in}}%
}

\endgroup

\section{Introduction}
\label{sec:introduction}
Generative-AI systems have moved rapidly from research prototypes to widely
used digital infrastructure. The 2026 Stanford AI Index reports that generative
AI reached an estimated $53\%$ population adoption within three years, while $70\%$ of surveyed organizations used generative AI in at least one business function in 2025~\cite{aiindex2026}. At consumer scale, ChatGPT reported $700$ million weekly active users in 2025; an analysis of $1.5$ million
privacy-preserving sampled conversations found that approximately three-quarters
concerned practical guidance, information seeking, or writing~\cite{openai2025usage}.
Such systems support search and question answering, translation, summarization,
writing assistance, conversational agents, and code-generation tools.

Much of this progress in language-focused generative AI has been enabled by
Transformer architectures. Prominent autoregressive and masked-language models,
including GPT-2, BERT, and RoBERTa, use Transformer-based representations to
map tokenized text into vector spaces from which subsequent contextual
computation and output generation proceed~\cite{vaswani2017attention,
radford2019language,devlin2019bert,liu2019roberta}. As these models become
embedded in products and research workflows, it becomes important to understand
not only the outputs they produce, but also the representational structures from
which those outputs arise.

Transformer models are commonly evaluated behaviorally through the tasks they
solve, the text they generate, and the benchmarks on which they succeed or fail.
Yet frozen Transformer input-embedding spaces may also be examined through their
responses to a controlled deterministic probe before contextual computation or
task-specific adaptation. Before contextual computation begins, a model-specific
tokenizer converts input text into discrete token IDs, and the learned
token-embedding table maps those IDs to vectors. These vectors occupy a
model-specific input-embedding space: the continuous geometric substrate from
which the model's subsequent computation begins.

Table~\ref{tab:acronyms} lists the acronyms and model identifiers used
throughout the paper, while Table~\ref{tab:notation} summarizes the principal
mathematical notation.

\begin{table}[!t]
    \centering
    \caption{Acronyms and model identifiers used throughout the paper.}
    \label{tab:acronyms}

    \footnotesize
    \setlength{\tabcolsep}{4pt}
    \renewcommand{\arraystretch}{1.16}

    \begin{tabularx}{\columnwidth}{
        @{}
        L
        C{0.34\columnwidth}
        @{}
    }
        \toprule
        \rowcolor{black!8}
        \textbf{Term}
            & \textbf{Acronym or identifier} \\
        \midrule

        Neurochaos Learning
            & NL \\
        Generalized L\"uroth Series
            & GLS \\
        Machine Learning
            & ML \\
        Deep Learning
            & DL \\
        Natural Language Processing
            & NLP \\
        Large Language Model
            & LLM \\

        \addlinespace[2pt]

        Generative Pre-trained Transformer
            & GPT \\
        Generative Pre-trained Transformer 2
            & GPT-2 \\
        Distilled GPT-2 model
            & \modelid{DistilGPT2} \\

        \addlinespace[2pt]

        Bidirectional Encoder Representations from Transformers
            & BERT \\
        Base uncased BERT model
            & \modelid{BERT-base-uncased} \\

        \addlinespace[2pt]

        Robustly Optimized BERT Pretraining Approach
            & RoBERTa \\
        Base RoBERTa model
            & \modelid{RoBERTa-base} \\

        \addlinespace[2pt]

        Nearest Neighbor
            & NN \\

        \bottomrule
    \end{tabularx}
\end{table}

\begin{table}[!t]
    \centering
    \caption{Principal mathematical notation used throughout the paper.}
    \label{tab:notation}

    \footnotesize
    \setlength{\tabcolsep}{5pt}
    \renewcommand{\arraystretch}{1.18}

    \begin{tabularx}{\columnwidth}{
        @{}
        C{0.19\columnwidth}
        L
        @{}
    }
        \toprule
        \rowcolor{black!8}
        \textbf{Symbol}
            & \textbf{Meaning} \\
        \midrule

        $\mathcal{V}$
            & Tokenizer vocabulary. \\
        $|\mathcal{V}|$
            & Number of tokens in the vocabulary. \\
        $d$
            & Token-embedding dimension. \\
        $\mathbf{E}$
            & Frozen token-embedding lookup table. \\
        $\mathbf{E}_{v,:}$
            & Embedding vector associated with token ID $v$. \\

        \addlinespace[2pt]

        $T$
            & Number of tokens in an input prompt. \\
        $\mathbf{X}$
            & Prompt-level token-embedding matrix. \\

        \addlinespace[2pt]

        $Q$
            & Initial neurochaotic neural activity. \\
        $B$
            & Discrimination threshold. \\
        $\epsilon$
            & Numerical matching tolerance. \\

        \bottomrule
    \end{tabularx}
\end{table}

For a transformer model with tokenizer vocabulary $\mathcal{V}$ and token
embedding dimension $d$, let
\begin{equation}
    \mathbf{E} \in \mathbb{R}^{|\mathcal{V}| \times d}
    \label{eq:token_embedding_matrix}
\end{equation}
denote the learned token-embedding lookup table. The row
$\mathbf{E}_{v,:}\in\mathbb{R}^{d}$ is the embedding associated with token
ID $v$. Thus, the rows of $\mathbf{E}$ form a finite set of learned token
embeddings in $\mathbb{R}^{d}$. For the pretrained models considered in this
study, $\mathbf{E}$ is held fixed throughout all experiments.

Given a prompt, the tokenizer produces a sequence of token IDs
$(i_1,i_2,\ldots,i_T)$, where $T$ denotes the tokenized sequence length.
The corresponding token-embedding matrix is obtained by lookup:
\begin{equation}
    \mathbf{X}
    =
    \begin{bmatrix}
        \mathbf{E}_{i_1,:} \\
        \mathbf{E}_{i_2,:} \\
        \vdots \\
        \mathbf{E}_{i_T,:}
    \end{bmatrix}
    \in \mathbb{R}^{T \times d}
    \label{eq:prompt_embedding_matrix}
\end{equation}
Here, $\mathbf{X}$ denotes the token-embedding lookup output before any architecture-specific positional, segment, normalization, or related input-processing operations. Each row of $\mathbf{X}$ contains the token embedding at one sequence position, while each column corresponds to one embedding dimension.
In this work, $\mathbf{X}$ contains only the vectors retrieved from the learned token embedding table; model-dependent positional or token-type embeddings and contextual hidden states are not included. The token vectors retrieved from the model's learned input-embedding table, before the addition of any model-dependent components or processing can be referred to as the \emph{input embedding}.
In the initial stage, the embeddings are non-contextualized: a given token ID retrieves the same learned vector wherever it occurs. The input-embedding space therefore forms the model's initial continuous organization of discrete tokens, that is, the geometry through which those tokens first enter its computational pipeline.

Although these input embeddings precede contextual computation, their geometry is neither empty nor arbitrary. Foundational work on distributed word representations demonstrated that learned embedding spaces can contain
meaningful semantic and syntactic regularities~\cite{mikolov2013efficient,pennington2014glove}. Subsequent work has also shown that language-model embedding spaces may also exhibit shared directional components, frequency-related organization, anisotropy, and representation degeneration~\cite{gao2019representation,bis2021too}. An experiment by Biś et al.~\cite{bis2021too} examined the non-contextualized embedding matrices of BERT, RoBERTa, and GPT-2 and identified systematic geometric structure before contextual processing begins. The findings establish frozen transformer input-embedding spaces as legitimate geometric objects of analysis, while also cautioning that their structure should not be conflated with the contextual hidden states generated by later transformer layers. Existing representation-analysis methods primarily characterize learned spaces through direct geometric measurements, cross-representation alignment, or learned diagnostic probes~\cite{ethayarajh2019contextual,kornblith2019similarity,rudman2022isoscore,belinkov2022probing}.
 

A complementary question therefore remains: can a frozen transformer input-embedding space be characterized by how it responds to a fixed deterministic nonlinear probe?

We approach this question through Neurochaos Learning (NL), a brain-inspired learning paradigm in which one-dimensional Generalized L\"uroth Series (GLS) chaotic maps generate neural firing trajectories from which descriptors such as Firing Rate, Firing Time, Energy, and Entropy are extracted~\cite{harikrishnan2020neurochaos,nb2022causality}. In established NL variants, these trajectory-derived descriptors are typically treated as features for classification or related predictive tasks and have been employed in various applications such as SARS-CoV-2 genome sequence classification~\cite{harikrishnan2022classification}, forest fire classification~\cite{pant2025advancing}, hypothetical protein classification~\cite{anusree2024hypothetical,sneha2023biologically} and for widely used Machine Learning (ML) datasets like Iris classification, Breast Cancer Wisconsin classification and many more~\cite{sethi2023neurochaos}.
Thus, the success or failure of the transformation is typically judged by downstream performance. In this work, instead of asking whether neurochaotic descriptors improve a classifier, we investigate the systematic response of a frozen transformer input-embedding space across neurochaotic probe configurations. From this point of view, the space is not only characterized by the coordinates it contains, but also by the response profile it presents to a controlled deterministic probe.

Guided by this response-based view, we introduce \emph{ChaosProbe}, a deterministic neurochaos-inspired framework for response-surface fingerprinting of frozen transformer input-embedding spaces. For each prompt-level input embedding matrix $\mathbf{X}$ defined in Equation~\eqref{eq:prompt_embedding_matrix}, ChaosProbe applies global min--max normalization and generates skew-tent trajectories governed by an initial neural activity $Q$ and discrimination threshold $B$. For every normalized embedding coordinate, trajectory-derived descriptors are computed using a matching tolerance $\epsilon$. The resulting descriptor responses are summarized through representation-level metrics over a fixed $Q$--$B$--$\epsilon$ grid, forming a response surface whose fixed-order vectorization yields a model-level response signature. The response signature characterizes the interaction between a frozen input-embedding space and a fixed deterministic neurochaotic probe.


The principal contributions of this work are as follows:

\begin{itemize}
    \item We introduce a deterministic neurochaos-inspired framework that
    extends trajectory-descriptor methods from classification-oriented feature
    extraction to response-surface fingerprinting of frozen transformer
    input-embedding spaces.

    \item We formulate a fixed $Q$--$B$--$\epsilon$ probing procedure that
    converts prompt-level input embedding matrices into 800-dimensional
    model-level response signatures.

    \item We show, within a four-model proof-of-concept setting, that the response signatures recover the expected GPT-2/DistilGPT2 and BERT-base-uncased/RoBERTa-base nearest-neighbor pairings under Pearson correlation, Spearman correlation, and cosine similarity. Euclidean distance recovers the same-family nearest neighbor for three of the four models.
    
    \item We assess prompt-level stability through bootstrap resampling and
    examine zero-variance and descriptor-collapse behavior to guard against
    trivial or predominantly degenerate response signatures.
\end{itemize}

\section{Background and Motivation}
\label{sec:background}

The motivation for ChaosProbe lies at the intersection of two lines of
research: the geometric analysis of learned representation spaces and the use
of deterministic chaotic dynamics to transform numerical inputs into
trajectory-derived descriptors. The former asks what structure a learned space
exhibits; the latter asks how an input is expressed through its interaction
with a controlled nonlinear system. ChaosProbe brings these perspectives
together, while restricting its object of study to frozen transformer
input-embedding spaces.

\subsection{Frozen Input-Embedding Spaces as Objects of Study}
\label{subsec:frozen_embedding_spaces}

A transformer's input-embedding matrix is learned during pretraining. Once the model is frozen,  the input-embedding matrix defines a fixed mapping from token IDs to vectors. These vectors constitute the model's non-contextualized lexical representation. Before the transformation layers operate, the same token identifier retrieves the same learned vector, independently of the surrounding sequence. These vectors only become context-dependent representations after being combined with any model-specific positional or token-type information and processed through the transformer layers.

Although input embeddings precede contextual computation, they are not arbitrary numerical initializations. Their organization emerges from the interaction among the training corpus, tokenization scheme, vocabulary, optimization process, and pretraining objective. Previous work has shown that transformer embedding spaces may exhibit directional concentration, frequency-related organization, and non-uniform utilization of their dimensions~\cite{bis2021too,rudman2022isoscore}. Such observations motivate the treatment of a frozen input-embedding space as a legitimate, although deliberately bounded, object of representation analysis.

An additional qualification concerns how the embedding space is sampled.
Each prompt samples a finite subset of rows from a model's full embedding
table, and different models may tokenize the same text differently. The
empirical object probed by ChaosProbe is therefore the collection of
prompt-level input-embedding matrices produced by the frozen embedding lookup
and model-specific tokenizer. The resulting response signature depends on the
embedding table, tokenizer, preprocessing procedure, and prompt distribution.
ChaosProbe consequently characterizes the input-embedding space as sampled by
the controlled prompt set rather than every vector in the complete vocabulary.

\subsection{Representation-Space Comparison and Probing}
\label{subsec:representation_comparison}

There are several complementary methodological traditions for studying neural representation spaces. Geometric approaches deal directly with similarity, alignment, anisotropy, dimensional use, and neighborhood structure. Instead, representational-similarity methods compare the relationship among activations across models or layers. For example, centered kernel alignment (CKA) measures the similarity between representation matrices and is invariant to some transformations of their coordinates~\cite{kornblith2019similarity}. Further work on isotropy has demonstrated that seemingly simple geometric properties can be substantially influenced by the manner in which they are mathematically defined and measured~\cite{rudman2022isoscore}.

Another method of analysis of representations involves probing classifiers, where an auxiliary model is trained to predict a particular linguistic or behavioral property from a representation \cite{belinkov2022probing}. Such methods lend themselves to tests of recoverability of specific information, but they need to be interpreted according to the capacity of the probe, the design of control tasks, and the distinction between encoded information and information actually used by the original model.

ChaosProbe takes another methodological direction. It does not learn an auxiliary classifier nor attempt to decode a linguistic property, or directly align embedding coordinates between models. This is important because the models considered here differ in vocabulary, tokenization, embedding dimensionality, and architectural objective, so a raw coordinate-wise comparison is ill-defined. Instead, ChaosProbe applies the same deterministic nonlinear probing procedure to each prompt-level input-embedding matrix and compares the resulting fixed-length response signatures.

\section{The ChaosProbe Framework}
\label{sec:chaosprobe_framework}

\subsection{Framework Overview}
\label{subsec:framework_overview}

ChaosProbe constructs a fixed-length signature for a frozen transformer model by
measuring how its prompt-level input embeddings respond to a controlled family
of deterministic neurochaotic probe configurations. This response-based
formulation maps models with different tokenizers, vocabulary sizes, sequence
lengths, and embedding dimensions into a common signature space.

Let $m$ denote a pretrained transformer model and let $p$ denote a prompt from a
common set of $P$ prompts. The model-specific tokenizer converts $p$ into a
sequence of token identifiers. Lookup in the frozen token-embedding matrix then
produces the prompt-level input-embedding matrix

\begin{equation}
    \mathbf{X}_{m,p}
    \in
    \mathbb{R}^{T_{m,p}\times d_m},
    \label{eq:model_prompt_embedding_matrix}
\end{equation}

where $T_{m,p}$ is the number of tokens retained for prompt $p$ and $d_m$ is the
embedding dimension of model $m$. Each row of $\mathbf{X}_{m,p}$ represents one
token position, while each column represents one embedding dimension.

\subsubsection{Prompt-Level Normalization}
\label{subsubsec:prompt-normalization}
Because embedding coordinates may occupy different numerical ranges across
models, each prompt-level matrix is globally min--max normalized over all of its
entries. Let
$x_{m,p}^{\min}=\min_{t,j}(\mathbf{X}_{m,p})_{t,j}$ and
$x_{m,p}^{\max}=\max_{t,j}(\mathbf{X}_{m,p})_{t,j}$. The normalized matrix is

\begin{equation}
    \widetilde{\mathbf{X}}_{m,p}
    =
    \frac{
        \mathbf{X}_{m,p}-x_{m,p}^{\min}
    }{
        x_{m,p}^{\max}-x_{m,p}^{\min}
    }.
    \label{eq:global_minmax_normalization}
\end{equation}

If $x_{m,p}^{\max}=x_{m,p}^{\min}$, we define
$\widetilde{\mathbf{X}}_{m,p}=\mathbf{0}$ to avoid division by zero.

Consequently,
$\widetilde{\mathbf{X}}_{m,p}\in
[0,1]^{T_{m,p}\times d_m}$.
Normalization preserves the shape of the prompt-level matrix while placing its
coordinates in the unit interval on which the chaotic map operates.

\subsubsection{Deterministic Neurochaotic Probing}

ChaosProbe evaluates each normalized prompt-level matrix over a fixed parameter
grid. The $c$th probe configuration is denoted by

\begin{equation}
    \boldsymbol{\theta}_{c}
    =
    \left(Q_c,B_c,\epsilon_c\right),
    \qquad
    c=1,\ldots,C,
    \label{eq:probe_configuration}
\end{equation}

where $Q_c$ is the initial neural activity, $B_c$ is the discrimination threshold, and $\epsilon_c$ is the matching tolerance. Evaluation is done via the Cartesian product of the parameter values listed in Table~\ref{tab:probe-parameter-grid}. The grid is fixed before model comparison and contains $10\times5\times2=100$ configurations.

\begin{table}[t]
    \centering
    \caption{Fixed parameter grid used to construct the response signatures.}
    \label{tab:probe-parameter-grid}
    \footnotesize
    \setlength{\tabcolsep}{3.8pt}
    \begin{tabular}{@{}lcc@{}}
        \toprule
        \textbf{Parameter}
        & \textbf{Values}
        & \textbf{Count} \\
        \midrule
        Initial activity, $Q$
        & $0.06,0.16,\ldots,0.96$
        & 10 \\
        Threshold, $B$
        & $0.06,0.16,\ldots,0.46$
        & 5 \\
        Tolerance, $\epsilon$
        & $0.06,0.16$
        & 2 \\
        \bottomrule
    \end{tabular}
\end{table}

\begin{figure*}[!t]
    \centering

    \begin{subfigure}[t]{0.475\textwidth}
        \centering
        \includegraphics[
            width=0.98\linewidth
        ]{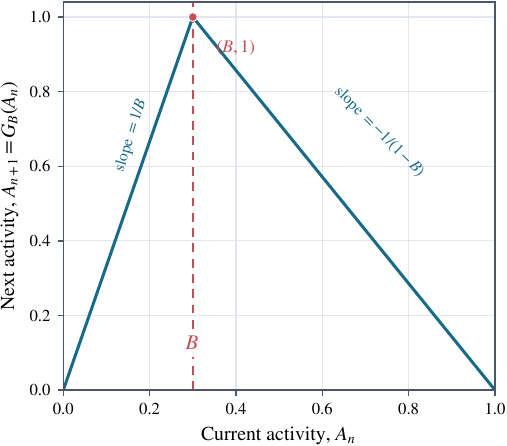}
        \caption{GLS skew-tent map.}
        \label{fig:gls-skew-tent-map}
    \end{subfigure}
    \hfill
    \begin{subfigure}[t]{0.475\textwidth}
        \centering
        \includegraphics[
            width=0.98\linewidth
        ]{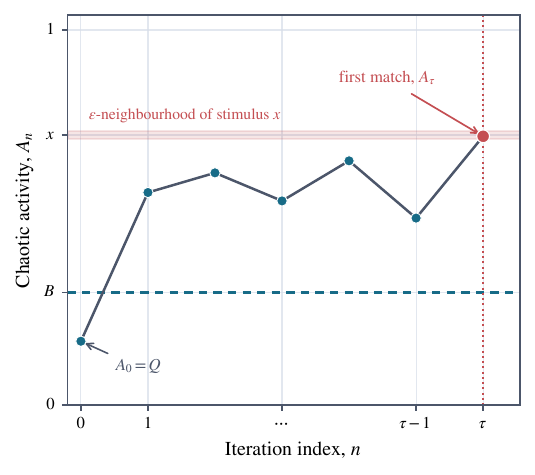}
        \caption{Trajectory and first $\epsilon_c$-match.}
        \label{fig:gls-first-match}
    \end{subfigure}

    \caption{Deterministic GLS probing mechanism used by ChaosProbe:
    (a) the skew-tent map parameterized by the discrimination threshold
    $B_c$, and (b) trajectory evolution from the initial neural activity
    $Q_c$ until the first $\epsilon_c$-match with stimulus $x$.}
    \label{fig:gls-dynamics}
\end{figure*}

For a given pair $(Q_c,B_c)$, the chaotic activity is initialized at
$A_0=Q_c$. Subsequent neural activity is generated by iterating the GLS
skew-tent map shown in
Fig.~\ref{fig:gls-dynamics}\subref{fig:gls-skew-tent-map}:
\begin{equation}
    A_{n+1}
    =
    G_{B_c}(A_n)
    =
    \begin{cases}
        \dfrac{A_n}{B_c},
        & 0\leq A_n < B_c, \\[8pt]
        \dfrac{1-A_n}{1-B_c},
        & B_c\leq A_n\leq 1.
    \end{cases}
    \label{eq:skew_tent_map}
\end{equation}

Every scalar coordinate
$x=(\widetilde{\mathbf{X}}_{m,p})_{t,j}$ is treated as a stimulus presented to the same deterministic trajectory. Let the set of matching trajectory indices be defined as
\begin{equation}
\mathcal{M}_L(x;\boldsymbol{\theta}_c)
=
\left\{
n\in\{0,\ldots,L-1\}:
\left|A_n-x\right|<\epsilon_c
\right\}.
\label{eq:matching_indices}
\end{equation} 
For a trajectory of fixed length $L$ = 1000, the first matching index is then
\begin{equation}
\tau_L(x;\boldsymbol{\theta}_c)
=
\begin{cases}
\min \mathcal{M}_L(x;\boldsymbol{\theta}_c),
& \mathcal{M}_L(x;\boldsymbol{\theta}_c)\neq\varnothing,\\[5pt]
L,
& \mathcal{M}_L(x;\boldsymbol{\theta}_c)=\varnothing.
\end{cases}
\label{eq:first_matching_index}
\end{equation}

Thus, $\tau_L=L$ indicates that no matching trajectory value was found within
the available trajectory.

The resulting trajectory and its first matching iteration are illustrated
in Fig.~\ref{fig:gls-dynamics}\subref{fig:gls-first-match}.

The trajectory segment used for descriptor calculation consists of the
trajectory values preceding the first matching index. Thus, when
$1\leq\tau_L(x;\boldsymbol{\theta}_c)\leq L$, the segment is
$(A_0,A_1,\ldots,A_{\tau_L(x;\boldsymbol{\theta}_c)-1})$. The matching
trajectory value itself is excluded. If the initial activity satisfies the
matching criterion, i.e., $\tau_L(x;\boldsymbol{\theta}_c)=0$, the segment is
defined as $(A_0)$ so that descriptor calculation remains nonempty. When no
match is found, $\tau_L(x;\boldsymbol{\theta}_c)=L$ and the complete
length-$L$ trajectory is used.

\subsubsection{Descriptor Channels}
\label{subsubsec:Descriptor-Channels}

Conventional Neurochaos feature extraction produces four trajectory-derived
descriptors: Firing Rate, Energy, Firing Time, and Entropy
\cite{sethi2023neurochaos}. ChaosProbe retains only Firing Rate and Entropy for
constructing the response signatures reported in this study.

Firing Rate is the fraction of the descriptor trajectory segment
$\mathcal{A}_{x,c}$ for which the neural activity exceeds the discrimination
threshold $B_c$. It is therefore a dimensionless threshold-occupancy proportion. Entropy is the binary Shannon entropy of the corresponding thresholded symbolic sequence, whose states record whether the neural activity exceeds $B_c$.

Let $\operatorname{FR}(x;\boldsymbol{\theta}_c)$ and
$\operatorname{ENT}(x;\boldsymbol{\theta}_c)$ denote the Firing Rate and
Entropy, respectively, obtained for scalar stimulus $x$ under parameter
configuration $\boldsymbol{\theta}_c$. Applying these descriptors
coordinatewise to $\widetilde{\mathbf{X}}_{m,p}$ gives

\begin{equation}
\begin{aligned}
    \left(
        \mathbf{D}^{\mathrm{FR}}_{m,p,c}
    \right)_{t,j}
    &=
    \operatorname{FR}\!\left(
        \left(
            \widetilde{\mathbf{X}}_{m,p}
        \right)_{t,j};
        \boldsymbol{\theta}_c
    \right), \\[4pt]
    \left(
        \mathbf{D}^{\mathrm{ENT}}_{m,p,c}
    \right)_{t,j}
    &=
    \operatorname{ENT}\!\left(
        \left(
            \widetilde{\mathbf{X}}_{m,p}
        \right)_{t,j};
        \boldsymbol{\theta}_c
    \right),
\end{aligned}
\label{eq:descriptor_channel_matrices}
\end{equation}

where $t=1,\ldots,T_{m,p}$ and $j=1,\ldots,d_m$. Thus,
$\mathbf{D}^{\mathrm{FR}}_{m,p,c}$ and
$\mathbf{D}^{\mathrm{ENT}}_{m,p,c}$ both belong to
$\mathbb{R}^{T_{m,p}\times d_m}$ and preserve the dimensions of the normalized
prompt-level embedding matrix. These matrices constitute the two descriptor
channels used in the subsequent representation-level analysis.

\subsubsection{Representation-Level Response Measures}
\label{subsubsec:representation-level-measures}

For every prompt, probe configuration, and retained descriptor channel,
ChaosProbe computes four representation-level measures:

\begin{equation}
\begin{split}
    \mathcal{K}
    =
    \{&
        \text{mean cosine similarity},
        \text{linear CKA},\\
      &
        \text{anisotropy},
        \text{normalized effective rank}
    \}.
\end{split}
\label{eq:representation_measure_set}
\end{equation}

\textit{Mean cosine similarity} compares each row of a descriptor-channel matrix with the corresponding row of the normalized input-embedding matrix and averages
the resulting cosine similarities over all retained token positions. It
therefore measures the average directional alignment between the original and
descriptor-transformed token representations. Consistent with the
implementation, a row with a zero norm contributes zero to the average.

\textit{Linear centered kernel alignment (CKA)} compares the centered relational
structure of the normalized input-embedding and descriptor-channel matrices
\cite{kornblith2019similarity}. Before computing linear CKA, both matrices are
centered by subtracting their respective column means. The feature-space
formulation is then applied to the centered matrices. A value of zero is
assigned when the normalization denominator vanishes.

\textit{Anisotropy} characterizes the internal directional concentration of a
descriptor-channel matrix. It is computed as the mean cosine similarity across
all distinct pairs of descriptor rows, excluding self-similarities
\cite{ethayarajh2019contextual}. Higher values indicate greater alignment among
the descriptor rows, whereas lower values indicate a more directionally
dispersed representation. Pairs containing a zero-norm row contribute zero.

\textit{Effective rank} characterizes the utilization of the singular-value spectrum of a matrix \cite{roy2007effective}. ChaosProbe uses a normalized form so that the measure is bounded independently of the maximum attainable matrix rank. Let
$\sigma_1,\ldots,\sigma_q$ denote the singular values of the uncentered
descriptor matrix $\mathbf{D}^{h}_{m,p,c}$, where
$q=\min(T_{m,p},d_m)$. The normalized effective rank is

\begin{equation}
\begin{aligned}
    p_i
    &=
    \frac{\sigma_i}{
        \sum_{j=1}^{q}\sigma_j
    }, \\[3pt]
    \operatorname{NER}
    \left(
        \mathbf{D}^{h}_{m,p,c}
    \right)
    &=
    \frac{1}{q}
    \exp\!\left(
        -\sum_{i=1}^{q}p_i\ln p_i
    \right).
\end{aligned}
\label{eq:normalized_effective_rank}
\end{equation}

The \textit{normalized effective rank} lies in $[0,1]$, with larger values indicating
broader utilization of the available singular directions. The implementation
assigns a value of zero to an all-zero descriptor matrix.

Let $\rho_{m,p,c,h,k}$ denote the scalar value obtained for model $m$, prompt
$p$, probe configuration $c$, descriptor channel $h$, and representation-level
measure $k$. Mean cosine similarity and linear CKA compare
$\widetilde{\mathbf{X}}_{m,p}$ with $\mathbf{D}^{h}_{m,p,c}$, whereas
anisotropy and normalized effective rank depend only on
$\mathbf{D}^{h}_{m,p,c}$.

Let $\mathcal{C}=\{\boldsymbol{\theta}_c\}_{c=1}^{C}$ denote the fixed set of
probe configurations and let
$\mathcal{H}=\{\mathrm{FR},\mathrm{ENT}\}$ denote the retained descriptor
channels. The model-level response for each configuration, channel, and measure
is obtained by averaging over the common set of $P$ prompts:

\begin{equation}
    \left(
        \mathbf{R}_{m}
    \right)_{c,h,k}
    =
    \frac{1}{P}
    \sum_{p=1}^{P}
    \rho_{m,p,c,h,k}.
    \label{eq:prompt_aggregated_response}
\end{equation}

The resulting response tensor satisfies

\begin{equation}
    \mathbf{R}_{m}
    \in
    \mathbb{R}^{
        |\mathcal{C}|
        \times
        |\mathcal{H}|
        \times
        |\mathcal{K}|
    }.
    \label{eq:model_response_tensor}
\end{equation}
\subsubsection{Response-Signature Construction}
\label{subsubsec:response-signature-construction}

The response surface is vectorized using an identical ordering for every model.
Parameter configurations are ordered by $Q$, then $B$, and then $\epsilon$.
Within each configuration, Firing Rate precedes Entropy, and the
representation-level measures follow the order given in
Eq.~\eqref{eq:representation_measure_set}. The model response signature is
therefore

\begin{equation}
    \mathbf{s}_{m}
    =
    \operatorname{vec}\!\left(\mathbf{R}_{m}\right)
    \in
    \mathbb{R}^{800}.
    \label{eq:model_response_signature}
\end{equation}

The dimensionality follows from the 100 parameter configurations, two
descriptor channels, and four representation-level measures:
$100\times2\times4=800$.

Before model signatures are compared, each signature dimension is standardized
across the $M$ evaluated models. For signature dimension $\ell$, let

\begin{equation}
\begin{aligned}
    \mu_{\ell}
    &=
    \frac{1}{M}
    \sum_{m=1}^{M}
    s_{m,\ell},
    \\
    \sigma_{\ell}
    &=
    \left[
        \frac{1}{M}
        \sum_{m=1}^{M}
        \left(
            s_{m,\ell}-\mu_{\ell}
        \right)^2
    \right]^{1/2},
    \\
    z_{m,\ell}
    &=
    \begin{cases}
        \dfrac{s_{m,\ell}-\mu_{\ell}}{\sigma_{\ell}},
        & \sigma_{\ell}>0, \\[8pt]
        0,
        & \sigma_{\ell}=0.
    \end{cases}
\end{aligned}
\label{eq:signature_standardization}
\end{equation}

The resulting vector $\mathbf{z}_m$ is the standardized comparison signature
for model $m$. Dimensions with zero variance across models are assigned zero
after standardization. Pearson correlation, Spearman correlation, cosine
similarity, and Euclidean distance are subsequently computed between the
standardized signatures.

The unstandardized signature $\mathbf{s}_m$ is algorithmically deterministic
conditional on fixed model and tokenizer revisions, software environment,
prompt set and ordering, preprocessing procedure, parameter grid, and
aggregation order.
The standardized signature $\mathbf{z}_m$, however, also depends on the set of
models included in the comparison because its mean and standard deviation are
estimated across that model set.

\subsubsection{Model-Level Signature Comparison}
\label{subsubsec:model-signature-comparison}

Model relationships are evaluated using the standardized signatures
$\mathbf{z}_m$. Pearson correlation measures linear correspondence between
signature dimensions, whereas Spearman correlation measures correspondence
between their rank orderings. Cosine similarity measures the angular agreement
between two standardized signature vectors, while Euclidean distance measures
their absolute separation. Higher Pearson, Spearman, and cosine values indicate
greater similarity; lower Euclidean values indicate greater proximity.

The cosine similarity used here operates on complete model-level signatures and
is distinct from the mean cosine similarity used as a representation-level
response measure in Section~\ref{subsubsec:representation-level-measures}.
The latter compares corresponding rows of a normalized prompt-level embedding
matrix and its descriptor-channel matrix.

For each comparison method, the nearest neighbor of a model is selected from
the remaining models. The neighbor with the highest value is selected for the
three similarity measures, whereas the neighbor with the lowest value is
selected for Euclidean distance. Nearest-neighbor family accuracy is the
fraction of evaluated models whose selected neighbor belongs to the same
preassigned model family. A model pair is considered mutually recovered when
each member selects the other as its nearest neighbor.
If an exact tie occurs, the implementation selects the first candidate in the
fixed model ordering.

\subsection{Experimental Setup}
\label{subsec:experimental-setup}

\subsubsection{Models and Prompt Set}
\label{subsubsec:models-prompts}

The study evaluates four pretrained transformer checkpoints: GPT-2,
DistilGPT2, BERT-base-uncased, and RoBERTa-base. GPT-2 and DistilGPT2 are
assigned to the autoregressive GPT-style family, whereas BERT-base-uncased and
RoBERTa-base are assigned to the masked-encoder family
\cite{radford2019language,devlin2019bert,liu2019roberta}. These family
assignments are specified before response-signature comparison and are used
only to evaluate broad nearest-neighbor family recovery.

A common set of 80 English-language prompts is used for all four models. The
prompt set contains 20 prompts from each of four categories: factual,
reasoning, mathematical, and conversational. The prompts are neutral and
non-adversarial and are used solely to sample each model's frozen
input-embedding space. Prompt answers, generated continuations, and task
accuracy are not evaluated.

\subsubsection{Tokenization and Preprocessing Settings}
\label{subsubsec:tokenization-preprocessing}

Each prompt is tokenized independently using the tokenizer associated with its
pretrained checkpoint. Sequence packing and padding are not applied, and the
short prompts do not require truncation. Token rows identified by the tokenizer
as special tokens are excluded before further processing.

The retained token IDs are mapped through the model's frozen input-embedding
table, and the resulting prompt-level matrices are converted to double-precision
floating-point arrays. Global min--max normalization is then applied separately
to each prompt-level matrix according to the procedure defined in
Section~\ref{subsubsec:prompt-normalization}.

\subsubsection{Correlation-Based Bootstrap Stability Analysis}
\label{subsubsec:bootstrap-stability}

Prompt-level stability is evaluated using 200 bootstrap resamples. In each bootstrap iteration, 80 prompt indices are sampled with replacement from the common prompt set. The same sampled indices are used for all models so that prompt correspondence is preserved.

Prompt aggregation, response-signature construction, and cross-model standardization are repeated for every bootstrap sample. Stability is evaluated for Pearson and Spearman comparisons using two criteria. First, the mean within-family similarity is compared with the mean between-family similarity. Second, mutual nearest-neighbor recovery is recorded separately for the GPT-2--DistilGPT2 and BERT-base-uncased--RoBERTa-base pairs.

The reported bootstrap statistics are the fractions of resamples for which within-family similarity exceeds between-family similarity and for which the expected mutual nearest-neighbor pairs are recovered.

\subsubsection{Evaluation Protocol}
\label{subsubsec:evaluation-protocol}

For each model-level comparison measure---Pearson correlation, Spearman correlation, cosine similarity, and Euclidean distance, we compute nearest-neighbor assignments, nearest-neighbor family accuracy, and the number of mutually recovered expected family pairs. Pearson and Spearman similarity matrices are visualized in Section~\ref{sec:results} and are additionally assessed using 200 paired prompt-bootstrap resamples. Zero-variance signature dimensions and descriptor-channel collapse rates are reported as signature-validity checks.

\begin{table}[t]
\centering
\caption{Interpretation criteria and thresholds for the bounded four-model evaluation.}
\label{tab:operational_criteria}
\renewcommand{\arraystretch}{1.15}
\setlength{\tabcolsep}{3pt}
\scriptsize
\begin{tabular}{p{0.34\columnwidth}p{0.55\columnwidth}}
\toprule
\textbf{Criterion} & \textbf{Interpretation threshold} \\
\midrule
Family-level accuracy
& At least \(0.75\) for each comparison method; \(1.00=4/4\) denotes full recovery \\

Mutual-pair recovery
& At least one expected mutual pair per comparison method; \(2/2\) denotes full recovery. \\

Bootstrap stability
& At least \(0.80\) of paired resamples for Pearson and Spearman only. \\

Signature variation
& A zero-variance signature-dimension fraction below \(0.20\). \\

Descriptor non-collapse
& An all-zero response fraction below \(0.20\) for every model.  \\   
\bottomrule
\end{tabular}
\end{table}

Table~\ref{tab:operational_criteria} provides interpretation thresholds for
the bounded four-model evaluation. The family-recovery thresholds are applied
to all four comparison measures, whereas the bootstrap threshold applies only
to Pearson and Spearman correlation. These thresholds are descriptive
interpretation guides for this fixed cohort; they are not combined into a
single aggregate pass--fail decision rule.

\section{Results}
\label{sec:results}
\subsection{Parameter-Response Variation Across the Probe Grid}
\label{subsec:parameter-response-results}

\begin{figure*}[t]
\centering
\begin{minipage}{0.48\linewidth}
    \centering
    \includegraphics[width=\linewidth]
    {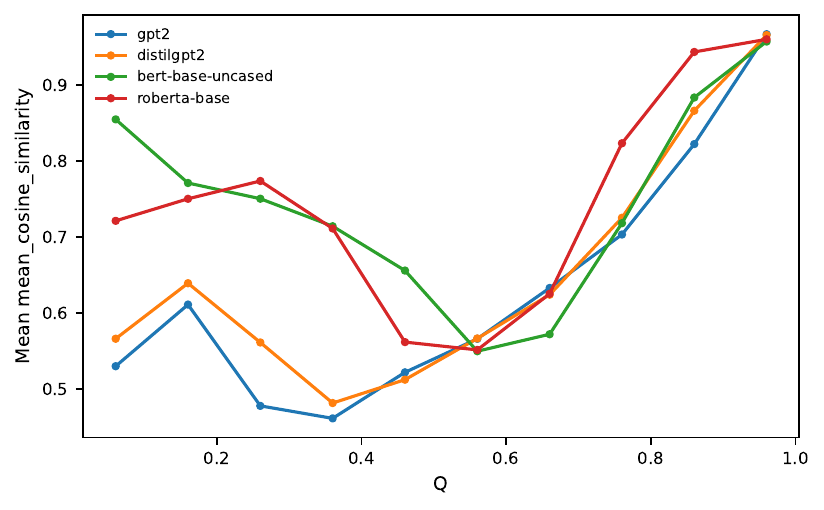}
    \small (a) Mean cosine similarity.
\end{minipage}\hfill
\begin{minipage}{0.48\linewidth}
    \centering
    \includegraphics[width=\linewidth]
    {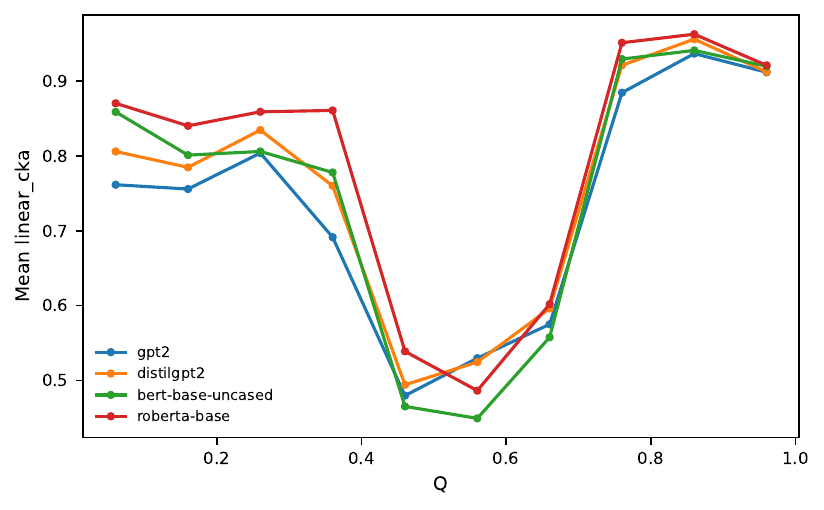}
    \small (b) Linear CKA.
\end{minipage}
\caption{Parameter-response variation across the initial-activity parameter
$Q$ for the four evaluated models: (a) mean cosine similarity and (b) linear
CKA. At each $Q$, each curve averages the prompt-level response over all 80
prompts, all five $B$ values, both matching tolerances, and both retained
descriptor channels. Both measures vary non-uniformly across the deterministic
probe grid, demonstrating that the response signatures contain
parameter-dependent structure rather than repetitions of an approximately
constant response.}
\label{fig:parameter_response}
\end{figure*}

Before comparing the final model-level signatures, we examine how the
representation-level responses vary across the probing grid.
Figure~\ref{fig:parameter_response} presents one-dimensional summaries of mean
cosine similarity and linear CKA as the initial neural activity $Q$ is varied.

Both measures exhibit pronounced variation across $Q$ rather than remaining
approximately constant. Mean cosine similarity generally decreases over the
lower-to-intermediate part of the grid and then increases toward the largest
$Q$ values. Linear CKA exhibits a related response pattern, with a marked
reduction at intermediate values of $Q$ followed by recovery at higher values.
The precise response magnitude and trajectory differ across the four models,
particularly in the lower and intermediate regions of the grid.

These results show that the probe parameters materially affect the measured
relationship between the input embeddings and their descriptor-derived
representations. Consequently, retaining responses from the complete fixed grid
captures parameter-dependent behavior that would be discarded if the
fingerprint were constructed from a single selected configuration.

\subsection{Model-Level Signature Similarity}
\label{subsec:signature-similarity-results}

\begin{figure*}[t]
\centering
\begin{minipage}{0.48\linewidth}
    \centering
    \includegraphics[width=\linewidth]
    {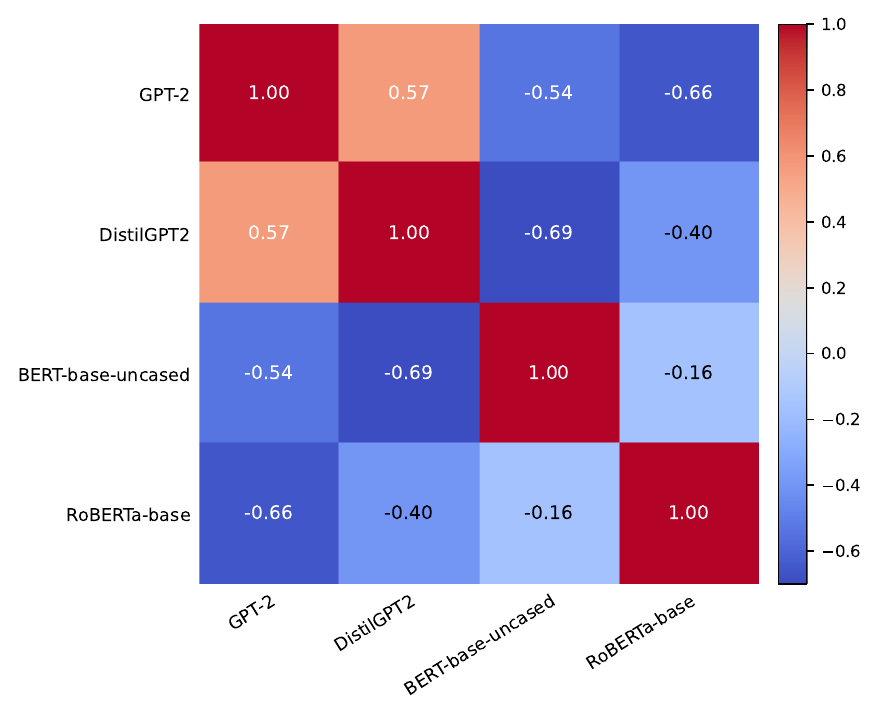}
    \small (a) Pearson correlation.
\end{minipage}\hfill
\begin{minipage}{0.48\linewidth}
    \centering
    \includegraphics[width=\linewidth]
    {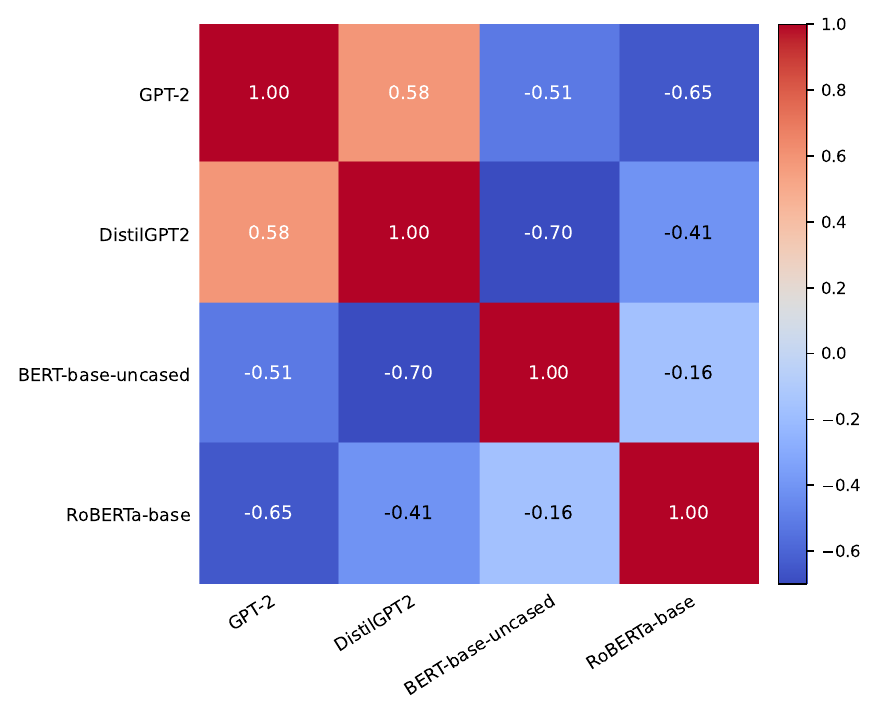}
    \small (b) Spearman correlation.
\end{minipage}
\caption{Pairwise comparison of the standardized 800-dimensional model
response signatures using (a) Pearson correlation and (b) Spearman
correlation. Under both measures, GPT-2 and DistilGPT2 are mutual nearest
neighbors, while BERT-base-uncased and RoBERTa-base are mutual nearest
neighbors. The negative BERT--RoBERTa correlations indicate relative
nearest-neighbor recovery rather than positive within-family agreement.}
\label{fig:signature_similarity_heatmaps}
\end{figure*}

Figure~\ref{fig:signature_similarity_heatmaps} compares the standardized
800-dimensional response signatures under Pearson and Spearman correlation.
Under Pearson correlation, GPT-2 and DistilGPT2 have a similarity of $0.57$.
This is the largest off-diagonal similarity for both models; their
cross-family correlations range from $-0.69$ to $-0.40$.

BERT-base-uncased and RoBERTa-base have a Pearson correlation of $-0.16$.
Although this value is negative, it is greater than their respective
correlations with the two GPT-style models, which range from $-0.69$ to
$-0.40$. Thus, BERT-base-uncased and RoBERTa-base are recovered as nearest
neighbors in relative correlation space, but the result does not indicate
positive correlation between their complete signatures.

Spearman correlation produces a closely related pattern. GPT-2 and DistilGPT2
have a rank correlation of $0.58$, while BERT-base-uncased and RoBERTa-base
again have a correlation of $-0.16$. The cross-family Spearman correlations
range from $-0.70$ to $-0.41$. Consequently, every model's highest
off-diagonal correlation is with the other member of its preassigned family
under both Pearson and Spearman comparison.

\subsection{Nearest-Neighbor Family Recovery}
\label{subsec:family-recovery-results}

\begin{table*}[t]
\centering
\caption{Nearest-neighbor family recovery across model-level signature
comparison methods.}
\label{tab:family_recovery}
\footnotesize
\setlength{\tabcolsep}{3.5pt}
\begin{tabular}{@{}lccc@{}}
\toprule
\textbf{Comparison}
& \textbf{Same-family assignments}
& \textbf{Family accuracy}
& \textbf{Mutual expected pairs} \\
\midrule
Pearson correlation  & $4/4$ & $1.00$ & $2/2$ \\
Spearman correlation & $4/4$ & $1.00$ & $2/2$ \\
Cosine similarity    & $4/4$ & $1.00$ & $2/2$ \\
Euclidean distance   & $3/4$ & $0.75$ & $1/2$ \\
\bottomrule
\end{tabular}
\end{table*}

Table~\ref{tab:family_recovery} summarizes nearest-neighbor recovery across the
four model-level comparison methods. Pearson correlation, Spearman correlation,
and cosine similarity each assign all four models to a nearest neighbor from
the same preassigned family, corresponding to a family-recovery accuracy of
$1.00$. Under each of these comparisons, both expected family pairs are
mutually recovered: GPT-2 selects DistilGPT2 and vice versa, while
BERT-base-uncased selects RoBERTa-base and vice versa.

Euclidean distance correctly assigns three of the four models to a same-family
nearest neighbor, giving an accuracy of $0.75$. It therefore recovers one of
the two expected family pairs mutually, while producing one cross-family
nearest-neighbor assignment. 

Thus, the expected family structure is recovered most consistently under correlation and cosine-based comparisons, whereas Euclidean distance reaches the minimum family-accuracy threshold with weaker mutual-pair recovery.

\subsection{Bootstrap Stability Under Prompt Resampling}
\label{subsec:bootstrap-results}

Table~\ref{tab:bootstrap_stability} reports the stability of the Pearson and
Spearman results across 200 paired bootstrap resamples of the 80 prompts. For
both comparison methods, the mean within-family similarity exceeded the mean
between-family similarity in all 200 resamples, producing an observed fraction
of $1.00$.

The two expected mutual nearest-neighbor pairs were also recovered
simultaneously in all 200 resamples under both Pearson and Spearman
correlation. Both correlation-based bootstrap statistics exceed the descriptive stability threshold of $0.80$ in Table~\ref{tab:operational_criteria}.

\begin{table}[t]
\centering
\caption{Prompt-resampling stability across 200 paired bootstrap resamples.}
\label{tab:bootstrap_stability}
\footnotesize
\setlength{\tabcolsep}{3.5pt}
\begin{tabular}{@{}lcc@{}}
\toprule
\textbf{Comparison}
& \textbf{Within $>$ between}
& \textbf{Both pairs} \\
\midrule
Pearson correlation
& $200/200\;(1.00)$
& $200/200\;(1.00)$ \\
Spearman correlation
& $200/200\;(1.00)$
& $200/200\;(1.00)$ \\
\bottomrule
\end{tabular}
\end{table}

These results indicate that the observed family-recovery pattern is stable to resampling and reweighting prompts within the evaluated prompt set. They do not, by themselves, establish stability for unseen prompts or different prompt
distributions.

Within the evaluated cohort, the correlation-based outcomes meet the
descriptive family-recovery and bootstrap-stability criteria in Table~\ref{tab:operational_criteria}. This conclusion is restricted to the evaluated checkpoints, prompt set, probe grid, descriptor channels, and comparison procedure. It should not be interpreted as evidence that ChaosProbe identifies arbitrary model families or generalizes to unseen architectures without further evaluation.

The signature-validity checks further indicate that the responses are not predominantly degenerate. Among the 800 unstandardized signature dimensions, 60 exhibited zero
cross-model variance ($60/800=0.075$), while the remaining 740 varied across
the evaluated cohort. At the prompt--configuration--channel level, a response
was classified as collapsed when its complete descriptor-response matrix was
zero. Collapse fractions ranged from $800/16{,}000=0.0500$ for RoBERTa-base
to $871/16{,}000=0.0544$ for GPT-2. Both quantities remain below the
interpretation thresholds in Table~\ref{tab:operational_criteria}, indicating
that constant dimensions and all-zero responses do not dominate the reported
signatures.

\section{Discussion}
\label{sec:discussion}

ChaosProbe represents each evaluated model through an 800-dimensional response
signature derived from its frozen input-embedding responses across a fixed
neurochaotic probing surface. Similarity between two signatures therefore
indicates similar responses under the sampled configurations and
representation-level measures; it does not imply equality of model parameters,
token embeddings, linguistic behavior, or downstream capability.

The recovered relationships may reflect several interacting properties,
including embedding-table geometry, tokenization and vocabulary construction,
prompt-specific token sequences, normalization, and probe-response behavior.
GPT-2 and DistilGPT2 have a direct teacher--student relationship~\cite{distilgpt2modelcard}, while BERT-base-uncased and RoBERTa-base share a masked-encoder lineage. However, the
present experiment does not isolate the contribution of these factors, nor does
it analyze contextual hidden states or transformer-layer computation.

The neurochaotic component provides deterministic trajectories and
trajectory-derived descriptors that are aggregated into a model-level response
surface. The non-uniform parameter responses support using the complete fixed
grid rather than one selected probe configuration. Nevertheless, this study
does not establish that chaotic trajectories are necessary for the observed
family recovery or superior to matched non-chaotic transformations. The result
should therefore be interpreted as a bounded proof of concept within the
evaluated four-model cohort.

\section{Limitations}
\label{sec:limitations}

The present study has several limitations that bound the interpretation of its
results.

\begin{itemize}[leftmargin=*]

    \item \textbf{Small model cohort.}
    The evaluation contains only four checkpoints arranged into two two-model
    families. A nearest-neighbor family accuracy of $1.00$ therefore represents
    four correct assignments, rather than evidence of performance across a broad
    population of models. Additional architectures, scales, and independently
    trained checkpoints are required before making general model-family
    identification claims.

    \item \textbf{Restricted representation and prompt scope.}
    Only frozen input-embedding matrices are analyzed; the results provide no
    evidence about contextual hidden states, attention patterns, or
    representation changes across transformer layers. Moreover, the evaluation
    uses 80 neutral English-language prompts. Bootstrap resampling assesses
    stability within this observed prompt set, but not generalization to
    independently collected prompts, longer contexts, multilingual inputs,
    specialized domains, or adversarial distributions.

    \item \textbf{Probe-grid sensitivity and limited ablation.}
    The study uses one fixed $Q$--$B$--$\epsilon$ grid and one trajectory length.
    Sensitivity to alternative parameter ranges, grid resolutions, trajectory
    lengths, and matching tolerances has not been evaluated. In addition, the
    final signatures combine Firing Rate and Entropy with four
    representation-level measures, and their individual contributions to family
    recovery have not been isolated.

    \item \textbf{Descriptive thresholds rather than calibrated decision rules.}
    The interpretation thresholds are practical safeguards for this bounded
    evaluation, rather than statistically calibrated universal cutoffs. They
    should not be interpreted as a formal hypothesis test or a general
    acceptance criterion for model-family recovery.

    \item \textbf{No downstream or safety evaluation.}
    The study does not evaluate text generation, task accuracy, robustness,
    jailbreak detection, prompt-injection detection, or other operational model
    capabilities. The reported evidence is restricted to
    representation-space fingerprinting.

\end{itemize}

\section{Conclusion and Future Work}
\label{sec:conclusion}

This work introduced ChaosProbe, a deterministic neurochaos-inspired framework
for constructing response-surface fingerprints from frozen transformer
input-embedding spaces. ChaosProbe applies a fixed grid of skew-tent trajectory
probes, transforms the resulting behavior into Firing Rate and Entropy descriptor
channels, and aggregates four representation-level measures into an
800-dimensional model response signature.

Within the evaluated four-model cohort, Pearson correlation, Spearman
correlation, and cosine similarity recover GPT-2--DistilGPT2 and
BERT-base-uncased--RoBERTa-base as mutual nearest-neighbor pairs. Euclidean
distance recovers three of the four same-family nearest-neighbor assignments and
one of the two expected mutual pairs. The correlation-based relationships remain
stable under 200 paired prompt-bootstrap resamples, while the signature-validity
diagnostics remain within their descriptive thresholds.

These findings support a bounded proof-of-concept conclusion: deterministic
neurochaotic response surfaces contain sufficient structure to recover the
expected relative relationships among the tested checkpoints. They do not
establish general model-family identification performance or chaos-specific
superiority.

Future work should evaluate larger balanced cohorts containing independently
trained checkpoints, varied model scales, and additional architecture families.
It should also develop an out-of-sample standardization protocol; compare the
neurochaotic construction with matched non-chaotic controls; assess channel,
measure, and probe-grid sensitivity; and test robustness across independently
collected prompt sets. Extending the analysis to contextual hidden states would
further reveal how these response signatures change across layers and after
adaptation such as fine-tuning or distillation.

\section*{Code Availability}
The implementation, experimental configuration, prompt set, and analysis scripts are available at
\url{https://github.com/kunal1704/ChaosProbe}.

\bibliography{references}
\bibliographystyle{unsrt}

\end{document}